\documentclass[conference]{ieeeconf}
\usepackage{cite}
\usepackage{amsmath,amssymb,amsfonts}
\usepackage{algorithm}
\usepackage{algorithmic}
\usepackage{amsmath}
\usepackage{graphicx}
\usepackage{textcomp}
\usepackage{arydshln}
\usepackage{xcolor}
\usepackage{todonotes}
\usepackage{array}
\usepackage{subcaption} % For subfigures
\usepackage{makecell}
\usepackage{booktabs}  
\usepackage{wasysym} 
\usepackage{placeins}  
\usepackage{hyperref}

\usepackage{tikz}
\usepackage{tcolorbox}
\usepackage{algorithm}
\usepackage{algorithmic}
\usepackage{amsmath}
\usepackage{xcolor}
\usepackage{graphicx}

\usepackage[utf8]{inputenc}
\usepackage{amsmath}
\usepackage{tcolorbox}
\tcbuselibrary{theorems}
\usepackage{algorithmic}
\usepackage{caption}

\newtcbtheorem[number within=section]{myalgorithm}{Algorithm}%
  {colback=gray!5, colframe=black, fonttitle=\bfseries, coltitle=black, rounded corners, boxrule=1pt}{algo}

\usetikzlibrary{shapes.geometric, arrows, positioning}

\def\BibTeX{{\rm B\kern-.05em{\sc i\kern-.025em b}\kern-.08em
    T\kern-.1667em\lower.7ex\hbox{E}\kern-.125emX}}
\begin{document}

\title{\textbf{Heterogeneous object manipulation on nonlinear soft surface through linear controller}}
\author{
    Pratik Ingle, Kasper Støy, Andres Faiña\\
    \small IT University, Denmark\\ 
    \small \{prin, ksty, anfv\}@itu.dk
}
\maketitle
\begin{abstract}
Manipulation surfaces indirectly control and reposition objects by actively modifying their shape or properties rather than directly gripping objects. These surfaces, equipped with dense actuator arrays, generate dynamic deformations. However, a high-density actuator array introduces considerable complexity due to increased degrees of freedom (DOF), complicating control tasks. High DOF restrict the implementation and utilization of manipulation surfaces in real-world applications as the maintenance and control of such systems exponentially increase with array/surface size. Learning-based control approaches may ease the control complexity, but they require extensive training samples and struggle to generalize for heterogeneous objects. 
In this study, we introduce a simple, precise and robust PID-based linear close-loop feedback control strategy for heterogeneous object manipulation on MANTA-RAY (Manipulation with Adaptive Non-rigid Textile Actuation with Reduced Actuation density). Our approach employs a geometric transformation-driven PID controller, directly mapping tilt angle control outputs(1D/2D) to actuator commands to eliminate the need for extensive black-box training. We validate the proposed method through simulations and experiments on a physical system, successfully manipulating objects with diverse geometries, weights and textures, including fragile objects like eggs and apples. The outcomes demonstrate that our approach is highly generalized and offers a practical and reliable solution for object manipulation on soft robotic manipulation, facilitating real-world implementation without prohibitive training demands.

\footnotetext{Funded by the European Union. Views and opinions expressed are however those of the author(s) only and do not necessarily reflect those of the European Union or the European Commission. Neither the European Union nor the granting authority can be held responsible for them.}

\end{abstract}

\section{Introduction}

\begin{figure*}[ht]
    \centering
    \includegraphics[width=0.97\linewidth]{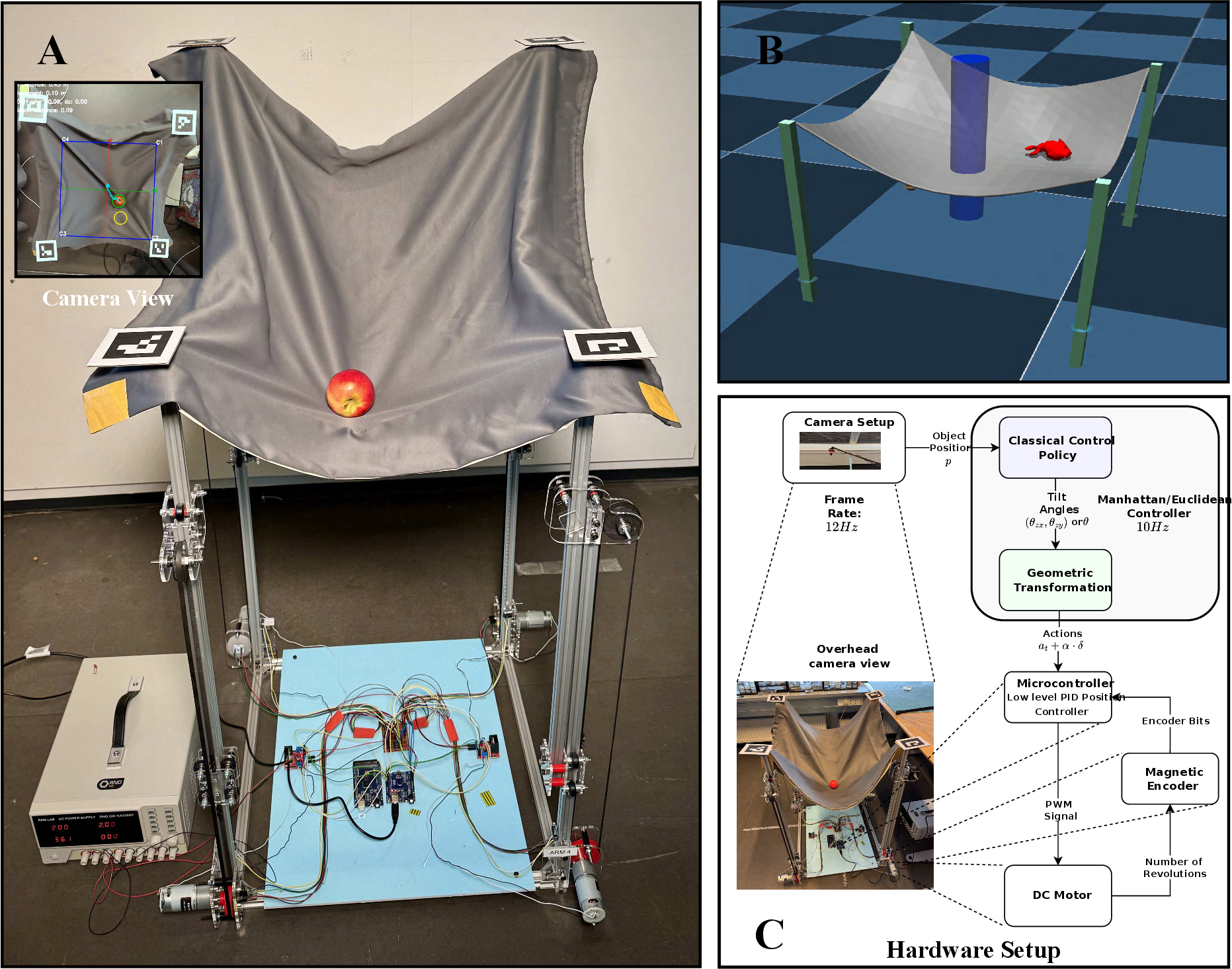}
    \caption{\textbf{A. MANTA-RAY:} A soft fabric is attached to four actuators positioned at the corners of a $0.5 \times 0.5$-meter aluminium frame. The system guides an apple in a circular path using an Arduino Uno microcontroller. The overhead camera view (top left) highlights the blue region of interest, with the object's trajectory over time shown in light blue, tracked with ArUco markers. \textbf{B. Simulation Setup:} MuJoCo physics engine with MANTA-RAY module size of $0.5 \times 0.5$-meter manipulating a Bunny(red) on the soft fabric, goal is to reach desire/target position(Blue) \textbf{C. Hardware Setup:} The hardware setup includes four actuators and an overhead camera for object tracking. An Arduino Uno microcontroller is employed to send PWM signals to the DC motors and to measure motor rotations using an AS500 magnetic encoder. Manhattan/Euclidean controller consist of clasical control policy That returns tilt angles which are used for geometrical transformation to get four action comparator actuator}
    \label{fig:ModuleSetup}
\end{figure*}

Objects come in a wide variety of shapes, sizes, and material properties, making it difficult for robotic manipulators to handle objects and generalize for heterogeneous objects. Traditional robotic grippers, while precise for structured tasks, often struggle with irregularly shaped or delicate objects. The gripper must apply careful force; too little and the object slip/drop; too much and a fragile item can crack. These limitations highlight the need for more adaptable manipulation strategies for mixed-object scenarios.

Robotic Manipulation Surfaces (RMS) have emerged as a promising approach for handling objects that vary widely in shape, size, and fragility – a scenario where traditional rigid manipulators often struggle. Instead of grasping objects directly, RMS use an array of actuators to deform or move a surface on which the object rests, thereby indirectly moving the object in combination or individually from friction, gravity and vibrations. The utilization of multiple actuators provides an extensive contact area, making them particularly effective for manipulating flat objects with larger surface areas and reducing the risk of damage. For example, wheel-based RMS \cite{parajuli2014actuator, uriarte2022methode} employ sets of omnidirectional wheels or rollers that transport objects by varying wheel speed, effective for high-throughput operations (e.g., sorting on assembly lines) and integrate well into logistics workflows. However, wheel-based methods are most effective on sufficiently large objects with flat, smooth bases; they cannot manipulate items significantly smaller than the wheel spacing. Another common design is piston-based RMS \cite{xue2024arraybot, johnson2023multifunctional}, which use dense grids of vertical pins or pistons that can individually lift and tilt beneath an object. Coordinated piston movements enable complex motions like translation, rotation, or even flipping of the object. However to reliably move objects of various sizes, these piston arrays must be extremely dense: if the target object is smaller than the gap between pistons, it may slip through, whereas if it is only supported by a single piston, motion is constrained to simple lifting. Consequently, most RMS implementations resort to high actuator density to ensure continuous support, which greatly increases the system’s complexity and cost. Beyond wheels and pistons, in literature several RMS techniques have been explored, such as vibrating plates \cite{georgilas2015cellular,zhou2016controlling} to shuffle objects. air flow levitation \cite{moon2006distributed} for frictionless transport, cilia-inspired micro-actuator arrays \cite{ataka20052d}, and deformable gel surfaces \cite{tadokoro2000distributed}. Each of these offers unique advantages, but they often address niche scenarios and still face trade-offs in payload, object compatibility, or control complexity.

To address these challenges, Ingle et al. \cite{11020841} proposed a novel system that drastically reduces the hardware and control complexity called MANTA-RAY: Manipulation with Adaptive Non-rigid Textile Actuation with Reduced Actuation densitY. MANTA-RAY uses soft manipulation surface with reduced actuator density – a system that achieves heterogeneous and fragile object manipulation over a large area with only a few actuators, described in section \ref{SMS_RAD}. 
It uses a flexible fabric sheet is attached to a small number of linear actuators forming a reconfigurable surface that can tilt and deform in order to move objects indirectly. By interconnecting the actuators with a continuous soft layer, the MANTA-RAY effectively uses the surface itself as a medium to transmit forces to the object, enabling even small or irregular items to be manipulated without falling through gaps. This approach dramatically lowers the actuator count and degrees of freedom \cite{wang2025surfacebasedmanipulation} compared to traditional RMS. The reduced hardware not only cuts down on system cost and weight, but also simplifies control since there are far fewer motors to coordinate. Despite this simplification, the MANTA-RAY retains the ability to handle heterogeneous and fragile objects. The soft surface provides gentle, adaptive support, demonstrated by successful manipulation of fragile items like eggs and fruits without damage. Notably, the system achieves an object-to-actuator size ratio as low as 0.01 (i.e., objects can be $100\times$ smaller than the spacing between actuators) – far beyond what previous manipulation surfaces could achieve. 

Despite these advantages, previous attempts at control policy with MANTA-RAY requires training reinforcement learning policy to manipulate objects(target reaching). Though it shows promises with objects like Box, Sphere and Disk, the controller does not generalise to any arbitrary object due to the non-linear behaviour of the fabric and sim-to-real transfer, as it requires a large number of training samples. Other strategies, eg ArrayBot \cite{xue2024arraybot} with  dense actuator array also uses RL policy for object manipulation and requires extensive training, making it harder for real-world deployment.
% This capability underscores the advantage of combining compliance with strategic hardware minimalism.

In this work we proposed a control strategy for the MANTA-RAY. Rather than resorting to complex non-linear controllers or black-box learning, we demonstrate that a classical linear controller can effectively control this soft, non-linear system when paired with an appropriate geometric transformation. In particular, we employ a PID-based close-loop feedback control scheme augmented with a geometric transformation. The controller operates in a low-dimensional action space of two/one dimensions. Instead of issuing separate commands to each actuator, it computes two high-level control outputs corresponding to the desired tilt angles of the surface in orthogonal directions. These two commands – conceptually akin to pitching the surface along X and Y axes – are then translated via geometric transformation into the four actions of actuators. By lowering the dimensionality of the control space from four to two, we significantly reduce the complexity of the control problem and ensure smoother, more coordinated actuator behavior. This approach effectively treats the MANTA-RAY as a platform whose shape can be linearly adjusted (within the small-angle regime) to move objects in desired directions, simplifying the otherwise highly non-linear dynamics of a fabric surface into a more tractable form. The PID controller continuously adjusts the tilt commands based on feedback – in our case, using visual tracking of the object’s position on the surface – to robustly guide the object to a target position. Such a closed-loop linear control with feedback achieves reliable manipulation without any learning phase, avoiding the heavy computational cost and uncertainty of RL while still compensating for unmodeled dynamics through error correction. Our method attains effective and stable manipulation on a soft non-linear surface that would otherwise be challenging to control. This represents a step toward more scalable and deployable MANTA-RAY, as fewer actuators and simpler controllers mean easier maintenance, lower cost, and better extensibility to larger surfaces or multi-agent setups.

\section{MANTA-RAY} \label{SMS_RAD}

MANTA-RAY system was introduced to achieve cost-effective manipulation of heterogeneous and fragile objects figure \ref{fig:ModuleSetup}. Having to rely on fewer actuators connected by a soft fabric layer covering large workspace areas makes the system work in low-action space and less complex to manage. Object manipulation on MANTA-RAY achieved through coordinated vertical displacement of the actuators, which induced controlled deformation in the fabric layer. These deformations generate localized slopes and tension variations on the surface, facilitating indirect movement of objects purely through altering the surface on which the object rests via gravity and frictional forces. The soft fabric layer is attached to actuators positioned at the corners of a $0.5 \times 0.5$-meter aluminium frame; based on application and workspace requirements, the soft fabric layer can be replaced, and the frame can be adapted to any $m \times n$-meter frame. Each actuator is capable of verticle movement of 0.5 meters, allowing the surface to form inclined planes or wave-like patterns. By adjusting the relative heights of the actuators, the system creates gradients that guide objects on the fabric. 

A primary method for manipulating an object on soft fabric involves inducing following motion behaviors:

    1. \textit{Rolling}: This motion is primarily governed by gravitational pull, friction and the surface’s inclination angle. Spherical or cylindrical objects roll when exposed to sufficient slope and minimal resistance from the fabric’s texture. 
    
    2. \textit{Sliding:}  Objects with a low center of gravity or higher surface contact (eg. disk) tend to slide down the slopes formed by the soft surface. The degree of sliding is influenced by the material properties of the fabric and the object’s weight and texture.
    
    3. \textit{Pulling:} Objects can also be displaced through indirect pulling mechanisms, where tension variations across the surface draw them along. Object position does not change with reference to the fabric. 

The tendency of an object to roll or slide depends largely on its shape. For example, a smooth, circular object is more likely to exhibit rolling motion, while a flat object with a low center of gravity will tend to slide rather than roll. The friction between the object and the soft fabric also influences these behaviors. Additionally, the motion can vary depending on whether the object is moving towards the edge or along the diagonal of the surface. The position control relies on precisely controlling the vertical positions of the actuators using a PID controller, which ensures smooth transitions between different surface configurations. The overhead camera tracks object positions in real-time using ArUco markers, providing feedback to adjust actuator movements accurately. The system’s ability to adapt the fabric’s shape dynamically allows it to handle objects of diverse shapes, sizes, and weights without direct contact between actuators and object. Furthermore, the manipulation efficiency depends significantly on the interplay between the object’s geometry and the soft surface’s characteristics, such as friction coefficient, elasticity, and surface tension.

\section{Methods}
The controller consists of two modules: 1. \textit{Controller:} If the object is too far, we tilt the surface more steeply in that direction until it moves closer. If it’s slightly off, we tilt gently. The controller continuously adjusts the tilt based on the difference between the current and desired object positions by using a linear PID controller. 2. \textit{Geometric transformation:} Tilt angles (1D/2D) from the controller mapped to four actuator commands, the amount each corner motor should raise or lower by considering the surface as a 2D plane passing through each actuator. This simple controller with two modules achieves robust heterogeneous object manipulation on the soft non-linear surface with only two-dimensional action space on a system with 4DOF with a low control frequency of 10Hz. 

Following sections \ref{subsec:controller} describe the controller and how linear PID controller in addition to geometric transformation is use to build two algorithms: Manhattan controller and Euclidean controller. Section \ref{subsec:simulation} and \ref{subsec:hardware} describe the simulation and hardware setup respectively used to evaluate the controllers in detail.

\subsection{Controller}\label{subsec:controller}

Four actuators of MANTA-RAY forms a single module. A module can have an arbitrary size $m \times n$-meters(mainly rectangular) capable of heterogeneous object movement on a workspace of size $m \times n$-meters. Four linear actuators control a single module and have four DOFs.

The controller works for a module of size $m \times n$. Take inputs as desired target position $\mathbf{p}_d = (x_d, y_d) \in \{ (x,y) \in \mathbb{R}^2 \mid |x| \le \frac{m}{2}, \; |y| \le \frac{n}{2} \}$, actuator base position $\{(x_i,y_i)\}_{i=1}^{4} = \{ (x,y) \in \mathbb{R}^2 \mid x = \pm \tfrac{m}{2},\, y = \pm \tfrac{n}{2} \}$, reference height of the system (actuator positions at resting phase) $z_0 = 1.5m$, and noise parameter $\alpha$ for irregular objects, algorithm 1. 

\resizebox{0.75\columnwidth}{!}{%
\begin{tcolorbox}[colback=gray!5, colframe=black, title={Algorithm 1: Object Manipulation via Manhattan Tilt}, rounded corners, boxrule=1pt]\label{manhattan_algo}
\begin{algorithmic}[1] 
    \REQUIRE Desired position $\mathbf{p}_d=(x_d,y_d)$, actuator base positions $\{(x_i,y_i)\}_{i=1}^{4}$, PID parameters for $x$ and $y$, control frequency $f$, tolerance $\epsilon$, reference height $z_0$, Noise amplitude $\alpha$
    \ENSURE Actuator commands $\{a_i\}_{i=1}^{4}$
    \STATE Initialize $\mathbf{p}_d$, $\{(x_i,y_i)\}_{i=1}^{4}$, configure PID controllers, and set $z_0$ (e.g., $z_0=1.5$)
    \STATE Compute control period: $dt \leftarrow 1/f$
    \WHILE{$\|\mathbf{p} - \mathbf{p}_d\| \ge \epsilon$}
        %%%%%%%%%%%%%%%%%%%%% Controller Block %%%%%%%%%%%%%%%%%%%%%
        \begin{tcolorbox}[colback=blue!5, colframe=black!100, title={Controller}, rounded corners, boxrule=0.5pt, left=1mm, right=1mm]
            \STATE Measure current object position $\mathbf{p} = (x,y)$
            \STATE Compute errors:
            \[
            e_x \leftarrow x_d - x,\quad e_y \leftarrow y_d - y
            \]
            \STATE Update PID controllers:
            \[
            \theta_{zx} \leftarrow \text{PID}_x(e_x, dt),\quad \theta_{zy} \leftarrow \text{PID}_y(e_y, dt)
            \]
        \end{tcolorbox}
        %%%%%%%%%%%%%%%%%%%%% Geometric Transformation Block %%%%%%%%%%%%%%%%%%%%%
        \begin{tcolorbox}[colback=green!5, colframe=black!70, title={Geometric Transformation}, rounded corners, boxrule=0.5pt, left=1mm, right=1mm]
            \STATE Compute normal vector components:
            \[ n_x = \sin(\theta_{zx}), \]
            \[ n_y = \sin(\theta_{zy}), \]
            \[ n_z = \cos(\theta_{zx}) \]
            \STATE Define plane using $P_0=(0,0,z_0)$:
            \[
            n_x\,x+n_y\,y+n_z\,(z-z_0)= 0
            \]
            \FOR{$i=1$ \textbf{to} $4$}
                \STATE Compute intersection for actuator $i$:
                \[
                z_i = \frac{z_0 \, n_z - n_x\, x_i - n_y\, y_i}{n_z}
                \]
                \STATE Set actuator command:
                \[
                a_i = (z_0 - z_i) + \alpha \,\delta_i, \quad \delta_i \sim U(-1, 1)
                \]
            \ENDFOR
        \end{tcolorbox}
        %%%%%%%%%%%%%%%%%%%%% Actuation and Update Block %%%%%%%%%%%%%%%%%%%%%
        \begin{tcolorbox}[colback=orange!5, colframe=black!70, title={Actuation and Update}, rounded corners, boxrule=0.5pt, left=1mm, right=1mm]
            \STATE Send command vector $\{a_i\}$ to actuators
            \STATE Update sensor measurements
        \end{tcolorbox}
    \ENDWHILE
\end{algorithmic}
\end{tcolorbox}%
}

The controller works at a control frequency of $f=10Hz$. At each time steps 
object position $\mathbf{p} = (x, y) \in \{ (x,y) \in \mathbb{R}^2 \mid |x| \le \frac{m}{2}, \; |y| \le \frac{n}{2} \}$ along with desired position $\mathbf{p}_d$ gives an errors $e_x, e_y$ along x and y- axis. These iterative errors, along with PID, return two control outputs $\theta_{zx}$ tilt along the X-axis and $\theta_{zy}$ tilt along the Y-axis from the Z-axis comprising the action space to only two dimensions, controller module, lines 4-6 of the algorithm 1 .

\resizebox{0.75\columnwidth}{!}{%
\begin{tcolorbox}[colback=gray!5, colframe=black, title={Algorithm 2: Object Manipulation via Euclidean Tilt}, rounded corners, boxrule=1pt]\label{eucledian_algo}
\begin{algorithmic}[1]
    \REQUIRE Desired position $\mathbf{p}_d=(x_d,y_d)$, actuator base positions $\{(x_i,y_i)\}_{i=1}^{4}$, PID parameters, control frequency $f$, tolerance $\epsilon$, reference height $z_0$, Noise amplitude $\alpha$
    \ENSURE Actuator commands $\{a_i\}_{i=1}^{4}$
    \STATE Initialize $\mathbf{p}_d$, $\{(x_i,y_i)\}_{i=1}^{4}$, configure PID controllers, and set $z_0$ (e.g., $z_0=1.5$)
    \STATE Compute control period: $dt \leftarrow 1/f$
    \WHILE{$\|\mathbf{p} - \mathbf{p}_d\| \ge \epsilon$}
    
        %%%%%%%%%%%%%%%%%%%%%%%%%%%%%%%%%%%%%%%%%%%%%%%%%%%
        %%%   Controller Block
        %%%%%%%%%%%%%%%%%%%%%%%%%%%%%%%%%%%%%%%%%%%%%%%%%%%
        \begin{tcolorbox}[colback=blue!5, colframe=black!70, title={Controller}, rounded corners, boxrule=0.5pt, left=1mm, right=1mm]
            \STATE Measure current object position $\mathbf{p} = (x,y)$
            \STATE Compute errors: 
            \[
            e_x = x_d - x,\quad e_y = y_d - y\]
            \[e=\sqrt{e_x^2+e_y^2} \]
            \STATE Compute tilt angle: 
            \[
            \theta \leftarrow \text{PID}(e, 1/f)
            \]
        \end{tcolorbox}
        
        %%%%%%%%%%%%%%%%%%%%%%%%%%%%%%%%%%%%%%%%%%%%%%%%%%%
        %%%   Inverse Kinematics Block
        %%%%%%%%%%%%%%%%%%%%%%%%%%%%%%%%%%%%%%%%%%%%%%%%%%%
        \begin{tcolorbox}[colback=green!5, colframe=black!70, title={Geometric Transformation}, rounded corners, boxrule=0.5pt, left=1mm, right=1mm]
            \STATE Compute normal vector components:
            \[
            n_x \;=\;\frac{e_x}{d}\,\sin(\theta), 
            \quad
            n_y \;=\;\frac{e_y}{d}\,\sin(\theta),\]
            \[\quad
            n_z \;=\;\cos(\theta),
            \]
            
            \[
            n \;=\;
            \begin{cases}
            \bigl(n_x,\;n_y,\;n_z\bigr), & d>0,\\
            (0,\,0,\,1), & d=0.
            \end{cases}
            \]
            \STATE Define Plane Using $P_0=(0,0,z_0)$, set
            \[
            n_x\,x+n_y\,y+n_z\,(z-z_0)=0.
            \]
            \FOR{$i=1$ \textbf{to} $4$}
                \STATE Compute actuator command:
                \[
                z_i = \frac{z_0 \, n_z - n_x\, x_i - n_y\, y_i}{n_z}
                \]
                \STATE Set actuator command:
                \[
                a_i = (z_0 - z_i) + \alpha \,\delta_i, \quad \delta_i \sim U(-1, 1)
                \]
            \ENDFOR
        \end{tcolorbox}
        
        %%%%%%%%%%%%%%%%%%%%%%%%%%%%%%%%%%%%%%%%%%%%%%%%%%%
        %%%   Actuation and Update Block
        %%%%%%%%%%%%%%%%%%%%%%%%%%%%%%%%%%%%%%%%%%%%%%%%%%%
        \begin{tcolorbox}[colback=orange!5, colframe=black!70, title={Actuation and Update}, rounded corners, boxrule=0.5pt, left=1mm, right=1mm]
            \STATE Send command vector $\{a_i\}$ to actuators.
            \STATE Advance step and sensor readings.
        \end{tcolorbox}
        
    \ENDWHILE
\end{algorithmic}
\end{tcolorbox}
}

Angles values return by the controller are use to project normal position vector $\vec{n} = (n_x, n_y, n_z) \in R^3$ as
\begin{equation}
    n_x = \sin(\theta_{zx}),
    n_y = \sin(\theta_{zy}),
    n_z = \cos(\theta_{zx})
\end{equation}

A plane $P$ can be represented by its point-normal form as $\vec{n} \cdot (\vec{r} - \vec{r_0}) = 0$
Where:
\begin{itemize}
    \item \( \vec{n} = (n_x, n_y, n_z) \) is the normal vector to the plane.
    \item \( \vec{r} = (x, y, z) \) is any point on the plane.
    \item \( \vec{r_0} = (x_0, y_0, z_0) = (0,0, 1.5) \) is a specific point on the plane, centrer of the system at reference height 1.5 meter.
\end{itemize}
Expanding this, we get:
\begin{equation} \label{z_val}
n_x (x - x_0) + n_y (y - y_0) + n_z (z - z_0) = 0
\end{equation}
We are interested in the $z$ value on the plane where the actuators are placed, corners of the module, as it represents the height of any point of the plane. With equation \ref{z_val} and actuator base postion $(x_i,y_i)$ we get the height of the actuators $z_i$ for actuator $i$ from 1 to 4. Geometric Transformation module, line 7-12 algorithm 1.
\begin{equation}
    z_i = \frac{z_0 \, n_z - n_x\, x_i - n_y\, y_i}{n_z} , \quad z_0= 1.5 meter
\end{equation}
This value gives the action as equation 

\begin{equation}
    a_i = (z_o - z_i) + \alpha \,\delta_i  \quad (i=0,1,2,3).
\end{equation}
Where $\alpha$ is the amplitude of the noise and $\delta \sim U(-1, 1)$, the uniform probability between  [-1,1]. Noise in the actuator command produces small waves on the soft fabric, making the system handle irregular objects more efficiently without getting stuck on the surface. Each actuator has a linear range of $[-0.25, 0.25]$ meters. Based on the actuator range the maximum angles the plane could tilt in X and Y axis $(\theta_{zx}, \theta_{zy})$ is $[-26^\circ,26^\circ]$

The controller keeps updating the two angle values $\theta_{zx}$ and $\theta_{zy}$ at the control frequency of $f$ and update four actuator positions $a_i$. This controller works with two dimensional work space of angle values. with similar PID and plan $P$ and slight changes in the control module we can reduce the controller space to one dimensional output, algorithm 2, Euclidean algorithm. In second approach instead of manhattan distance $(e_x, e_y)$ components of target and object distance we used euclidean distance $d=\sqrt{e_x^2+e_y^2}$. PID return an angle $\theta$ from the eucledian distance $d$ and control frequency $f$, controller module algorithm 2. Unline the manhattan algorithm, components of the normal vector $n_x, n_y,n_z$ can be calculated by considering the $\theta$ from z-axis in the direction of the object-target vector as 

\begin{equation}\label{normal_eu}
    n=\begin{cases} \left(\dfrac{e_x}{d}\sin(\theta),\; \dfrac{e_y}{d}\sin(\theta),\; \cos(\theta)\right), & \text{if } d>0, \\
    (0,0,1), & \text{if } d=0.
    \end{cases}
\end{equation}
When target and object position in $R^2$ not the same, $d>0$ then the $\dfrac{e_x}{d} \text{and} \dfrac{e_y}{d}$ defines the normalized object-target horizontal direction in XY-plane, when $d=0$ normal vector points straingt along the z-axis. Action values are calculated similarly as before by using equation \ref{normal_eu} normal vector $(n_x,n_y,n_z)$ and equation \ref{z_val}, see geometric transformation module in algorithm 2. 

\subsection{Simulation Setup:}\label{subsec:simulation}
MuJoCo \cite{todorov2012mujoco} physics-based simulator used to test the controller in the simulation environment. \textit{flexcomp} and \textit{composite} elements from MuJoCo facilitate the creation of deformable objects. This element allows to the define flexible components, such as cloth or soft bodies, figure \ref{fig:ModuleSetup}B Simulation facilitate positional actuators of range [-0.25, 0.25] meter, connected by a soft fabric of size $0.5 \times 0.5$-meter to make it closely mimic the hardware system and increment at the frequency of 1000Hz. 

\subsection{Hardware Setup:}\label{subsec:hardware}
% \todo{add hardware and simulation setup figure}
The hardware setup consists of a 0.5 × 0.5-meter aluminum frame supporting four vertically mounted linear actuators, positioned at each corner, to facilitate soft surface-based object manipulation, figure \ref{fig:ModuleSetup}C. Each actuator has a 0.5-meter vertical range and is driven by an RS Pro 834-7590 (12V, 120 RPM) DC geared motor, which converts rotational motion into linear displacement using a pulley and belt mechanism. The actuators are mounted on a V-slot rail system equipped with rolling wheels to ensure smooth and precise movement. A soft manipulation surface, composed of 100\% polyester fabric (0.6 × 0.6 meters), is attached to the actuators and provides a flexible interface for object handling through rolling, sliding, and pulling motions. The system is controlled by an Arduino Uno microcontroller, which sends PWM signals to an H-bridge motor driver, with real-time position feedback obtained via AS500 magnetic encoders (12-bit resolution). Object tracking is achieved using an overhead camera operating at 12 Hz, with ArUco markers placed at the corners of the manipulation region to enable accurate position estimation. The modular design allows for adjustments in actuator spacing, fabric properties, and integration of additional modules to scale the system for larger manipulation tasks.

\section{Results}

We evaluated controller performance in the simulation primarily by success rate, defined as the percentage of trials in which the object successfully reached the target without falling off the surface. The total manipulation workspace of $0.5 \times 0.5$-meter in simulation is divided into 400 small subsections of each size $0.025 \times 0.025$-meter. Success on each subsection is a percentage of 10 trials. where each  trial consisted a randomly assigned desired/target positions $p_d$ within the subsection and a random object position $p$ anywhere on the fabric with uniform probability. A trial was deemed successful if the object’s final position fell within a defined tolerance $\epsilon$ meters of the desired location $p_d$ by the end of the motion of runtime $T $ without toppling or leaving the surface. Success rate is assign based on the percentage of these 10 trials.A success rate across 400 subareas produces a heatmap of the fabric. 
The manipulation capability on RMS can be characterized into three distinct regions:

	\textit{1.	Success Region:} Areas where object manipulation is highly accurate, consistent, and repeatable.
    
	\textit{2.	Uncertain Region:} Areas where object manipulation is feasible but inconsistent and non-repeatable.
    
	\textit{3.	Failure Region:} Areas where object manipulation is infeasible, exhibiting a significantly high probability of failure.

A control policy is evaluated based on these three regions. The larger the success region the better the control policy. The performance of both controllers is compared based on the success rate heat map in the simulation.

Controller efficiency is been tested with two different kinds of experiments in the simulation and on the hardware: 

1 \textit{Euclidean vs Manhattan Control:} heatmaps representing simulation-derived success rates is utilized to comparatively evaluate controller efficiency.

2. \textit{Target reaching on the hardware:} An object is placed at the random starting position near center on the fabric and and aim to reach a random desired target position on the fabric without falling out of the surface. 

% 2. \textit{Path Following:} An object is tasked to follow a predefined trajectory on the surface without falling out of the surface. 

All the experiments are performed with heterogeneous objects with varying sizes, shapes and weights as described in the table \ref{tab:objects}. PID parameter of the Manhattan $(\text{same for }PID_x\text{ and }PID_y (k_P=30.9, k_i=26.78, k_d=15.45))$ and Euclidean $(PID (k_P=86.72, k_i=27.91, k_d=10.19))$ controller are kept same for all objects only the noise amplitude $\alpha$ varies. Integral windup is used for all PID controllers.

\begin{table}[ht ]
\centering
\begin{tabular}{l l l l}
\Xhline{3\arrayrulewidth} % Thicker top line
\addlinespace[2pt] % Add space between top line and header
\textbf{Object} & \textbf{Weight (g)} & \textbf{Size (cm)} & \textbf{Fabrication Method} \\ 
\addlinespace[2pt]
\Xhline{1\arrayrulewidth} % Regular line between header and content
\addlinespace[2pt]
Sphere    & 12.5  & \diameter: 5.1           & Molding        \\ 
Cube      & 66    & 5 x 5 x 5               & 3D Printing    \\ 
Disk      & 3.8   & \diameter: 4, Thickness: 0.4 & 3D Printing    \\ 
Apple     & 131.4   & \diameter: 5--7.2        & Supermarket \\ 
Cylinder  & 25.7  & \diameter: 3.3, Height: 4.5   & Molding        \\ 
Egg       & 66.7  & \diameter: 4.1--6.2         & Supermarket      \\ 
Bunny     & 50.3  &  Height: 9, length: 9      & 3D Printing \\
\addlinespace[2pt] % Add space between the last row and bottom line
\Xhline{3\arrayrulewidth} % Thicker bottom line
\end{tabular}
\caption{Details of objects: weight, size, and fabrication method.}
\label{tab:objects}
\end{table}

\subsection{Euclidean vs Manhattan Control}\label{e_v_m}
A sphere is utilized to generate a success grid on the simulation surface using Euclidean and Manhattan control policies. For both policies, the noise amplitude $\alpha$ is maintained at zero, trials are executed for a runtime of 10 seconds, and controllers operate at a 10 Hz control frequency.

\begin{figure*}[!t]
    \centering
    % First row with three columns
    \begin{subfigure}{0.28\textwidth}
        \centering
        \includegraphics[width=\textwidth]{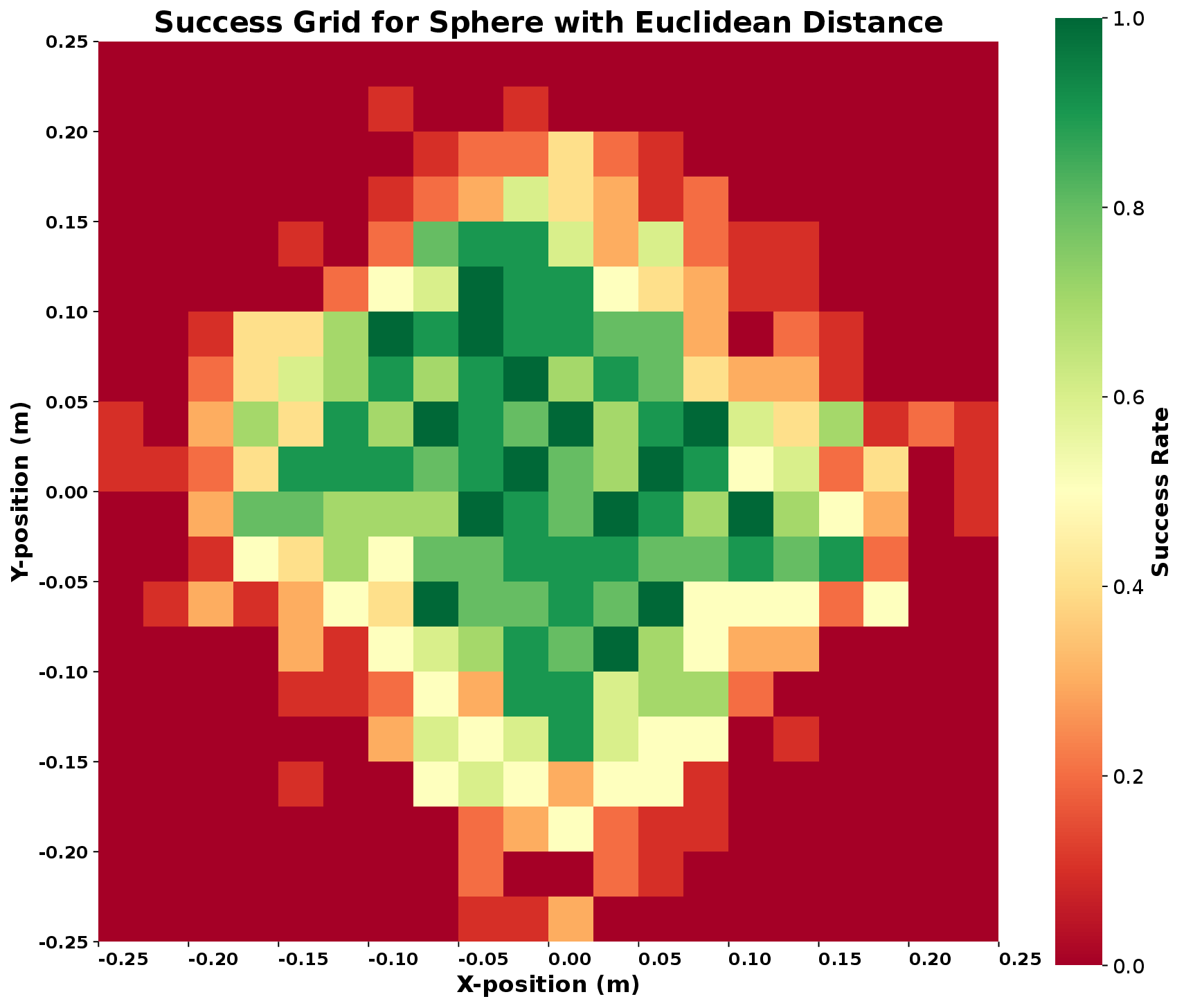} 
        \caption{Euclidean controller}
        \label{fig:ecl_sphere}
    \end{subfigure}
    \hfill
    \begin{subfigure}{0.28\textwidth}
        \centering
        \includegraphics[width=\textwidth]{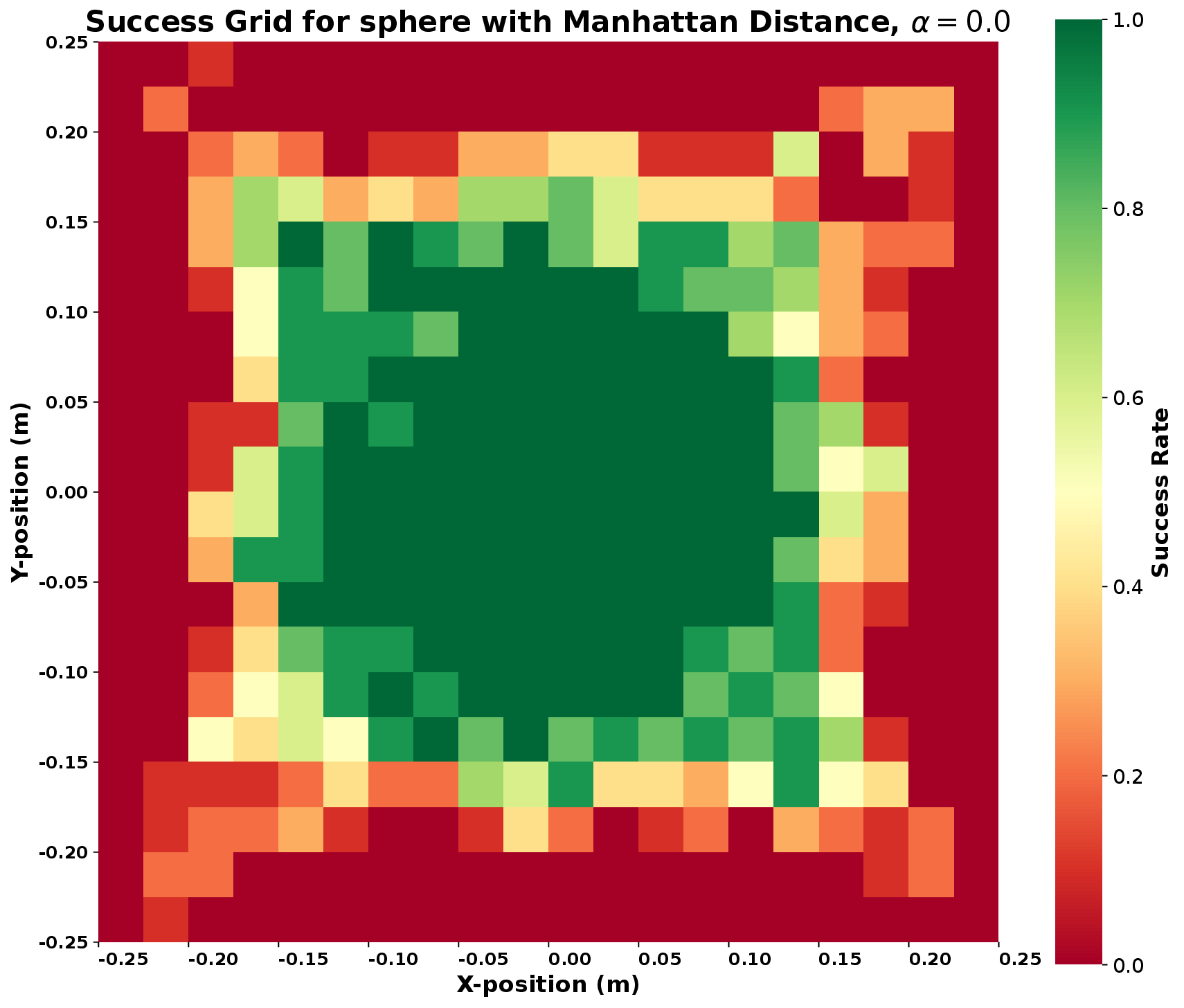} 
        \caption{Manhattan Controller}
        \label{fig:man_sphere}
    \end{subfigure}
    \hfill
    \begin{subfigure}{0.28\textwidth}
        \centering
        \includegraphics[width=\textwidth]{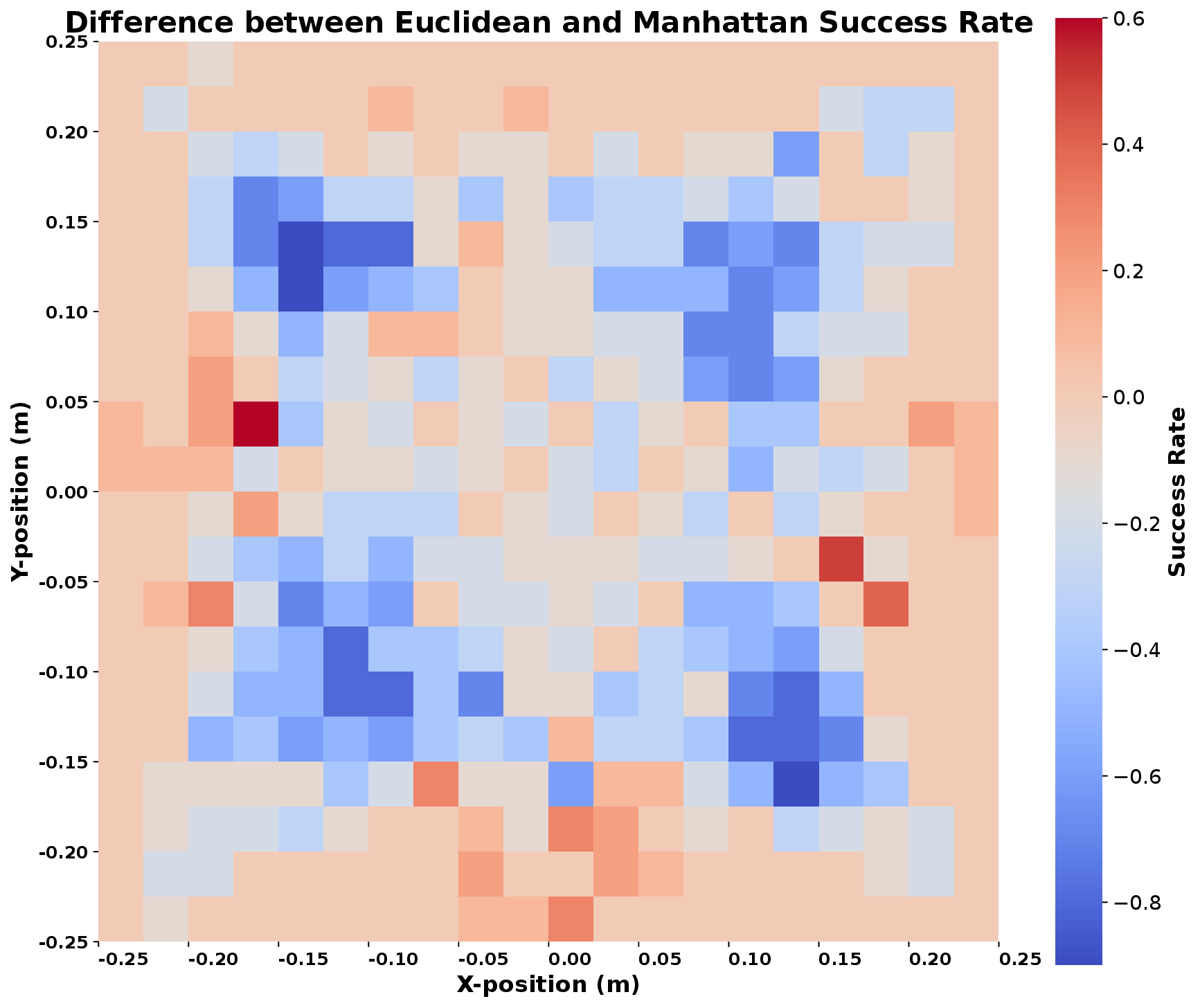} 
        \caption{Euclidean - Manhattan}
        \label{fig:diff_man_eucl}
    \end{subfigure}
    \hfill
    \begin{subfigure}{0.28\textwidth}
        \centering
        \includegraphics[width=\textwidth]{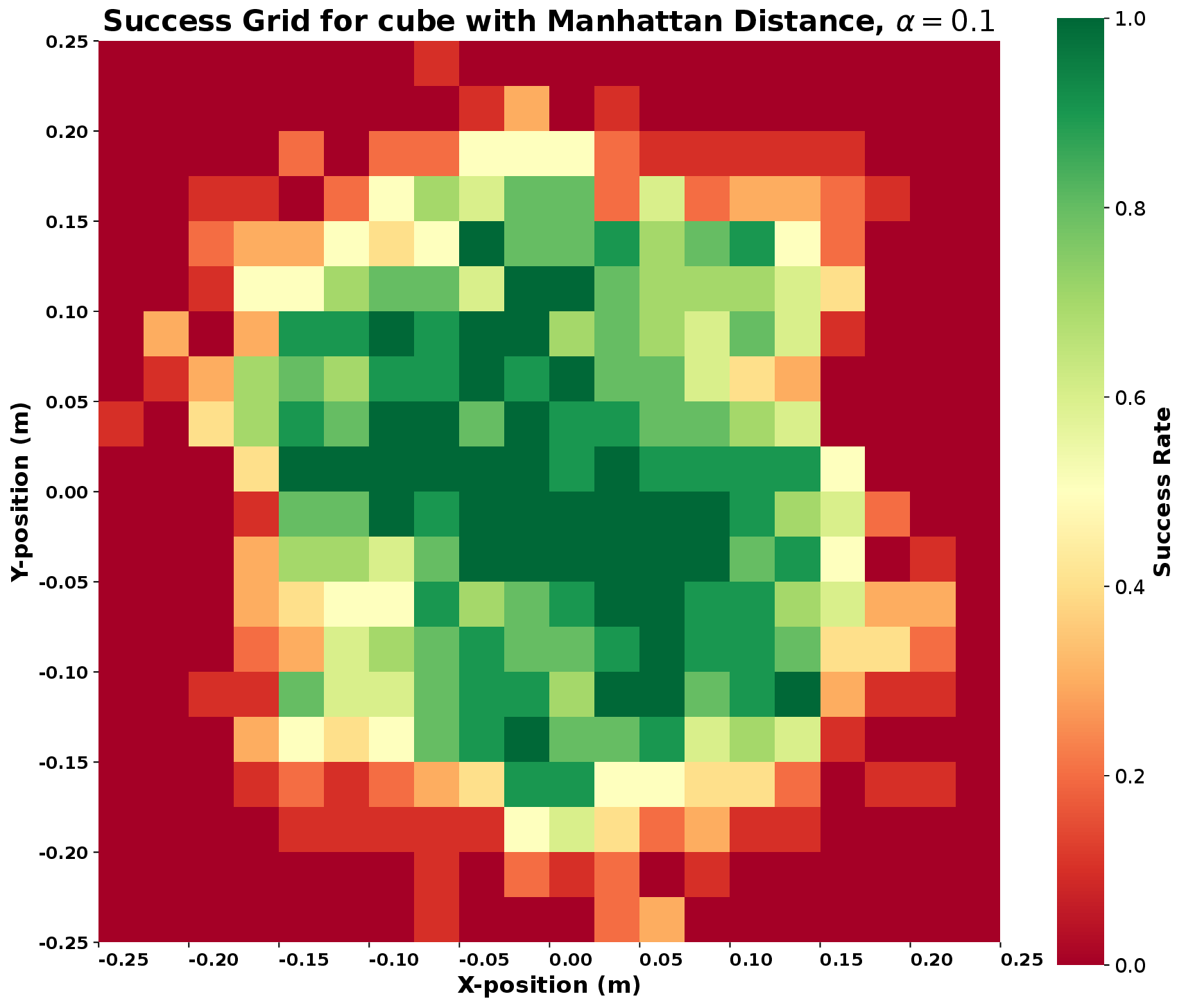} 
        \caption{Cube $\alpha = 0.1$}
        \label{fig:man_cube}
    \end{subfigure}
    \hfill
    \begin{subfigure}{0.28\textwidth}
        \centering
        \includegraphics[width=\textwidth]{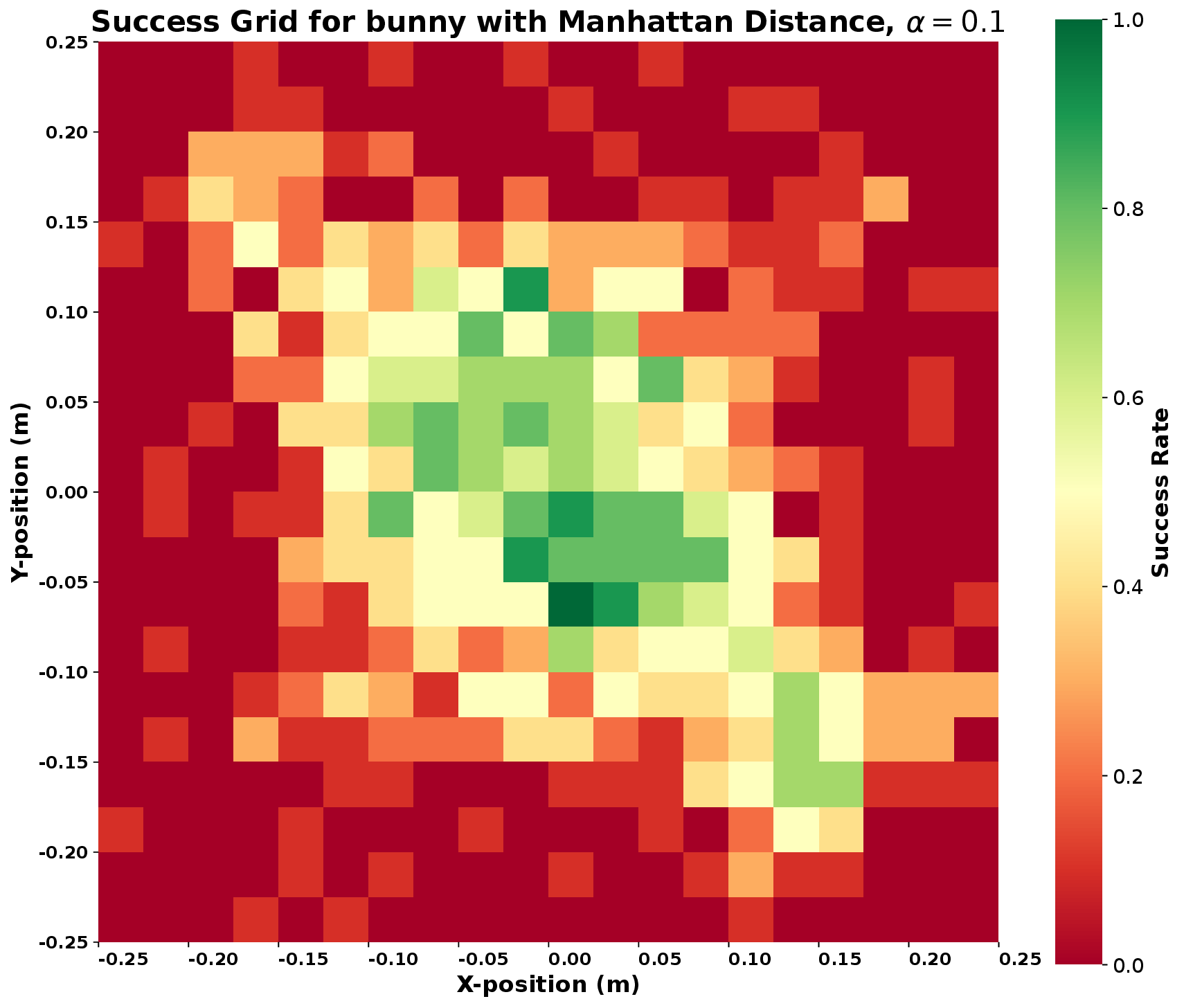} 
        \caption{Bunny, $\alpha = 0.1$}
        \label{fig:bunny_v0.1}
    \end{subfigure}
    \hfill
    \begin{subfigure}{0.28\textwidth}
        \centering
        \includegraphics[width=\textwidth]{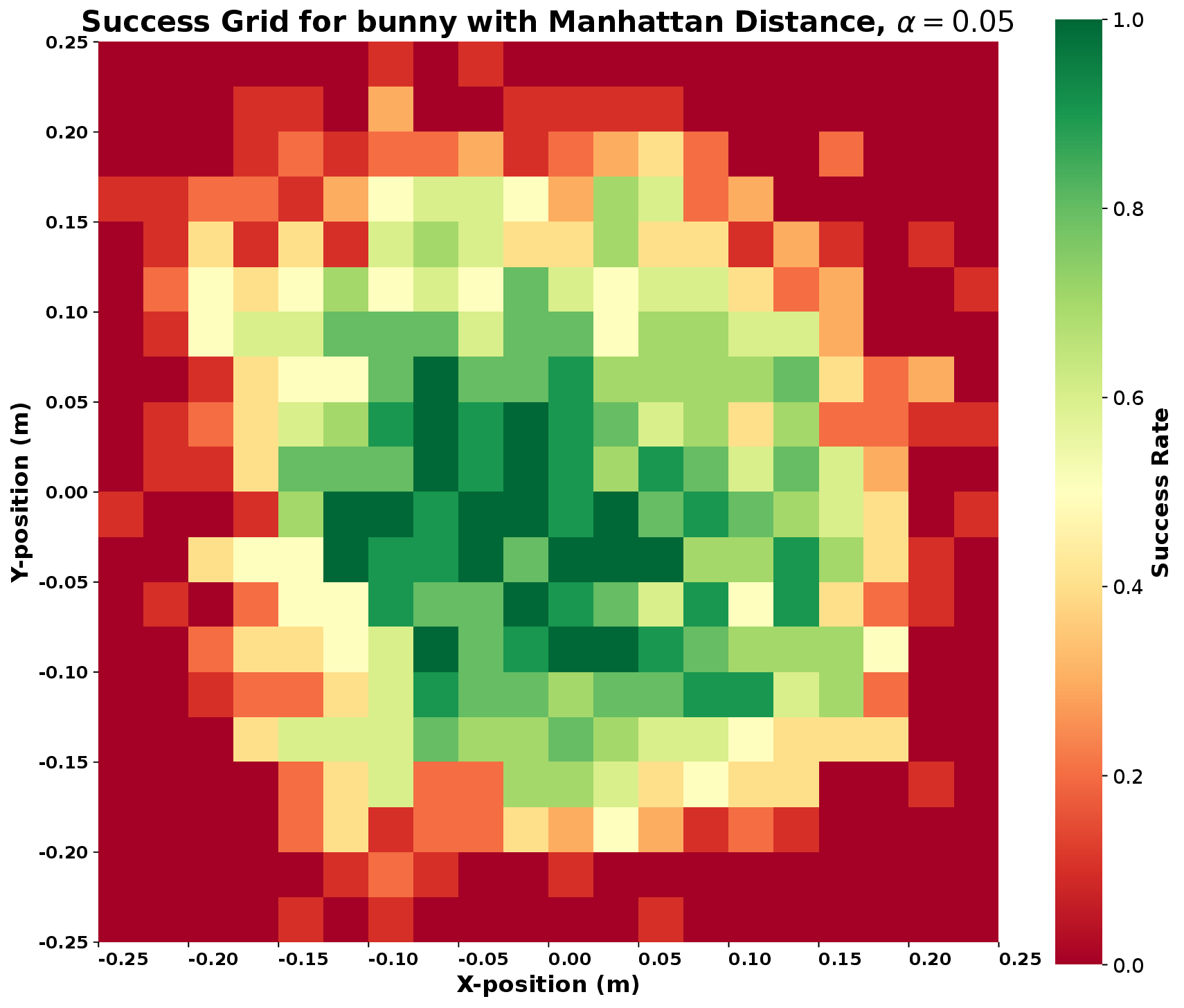} 
        \caption{Bunny, $\alpha = 0.05$}
        \label{fig:bunny_v0.05}
    \end{subfigure}
    
    \caption{Manhattan (b) vs Euclidean (a) controller for Sphere ($\alpha =0.0$) in simulation. Figure d, e and f shows success rate with the Manhattan controller for different objects: Cube (d), Bunny(e, f)}
    \label{fig:policy_com}
    
\end{figure*}

In Figure \ref{fig:ecl_sphere} illustrates the success rates achieved with the Euclidean control policy. A central success region emerges clearly, exhibiting 100\% effectiveness for some parts, surrounded by an uncertain region forming a diamond shape with moderate success rates, decreasing towards the fabric edges. The edges fall predominantly within the uncertain region, while corners remain entirely unreachable under Euclidean control. Conversely, the Manhattan control policy, figure \ref{fig:man_sphere} demonstrates a central region of 100\% success rate for all subregions, surrounded by a square-shaped uncertain region. Notably, this policy extends reachability to the corners, which were inaccessible using Euclidean control, although edges and borders remain challenging. This discrepancy between the two policies is depicted in Figure \ref{fig:diff_man_eucl}. The total Euclidean success rate on the overall surface is 108.20, while the Manhattan policy achieves 171.30, representing a 36\% improvement and enhanced corner accessibility. Success rate of Manhattan controller with different objects cube (\ref{fig:man_cube}), and bunny(\ref{fig:bunny_v0.1}, \ref{fig:bunny_v0.05}), also the effect of noise amplitude $\alpha$ (0.1 and 0.05) can be seen with Bunny.

\subsection{Target Reaching:}
\begin{figure*}[t]
    % \centering
    % First row with three columns
    \begin{subfigure}[b]{0.23\textwidth}
        \centering
        \includegraphics[width=\textwidth]{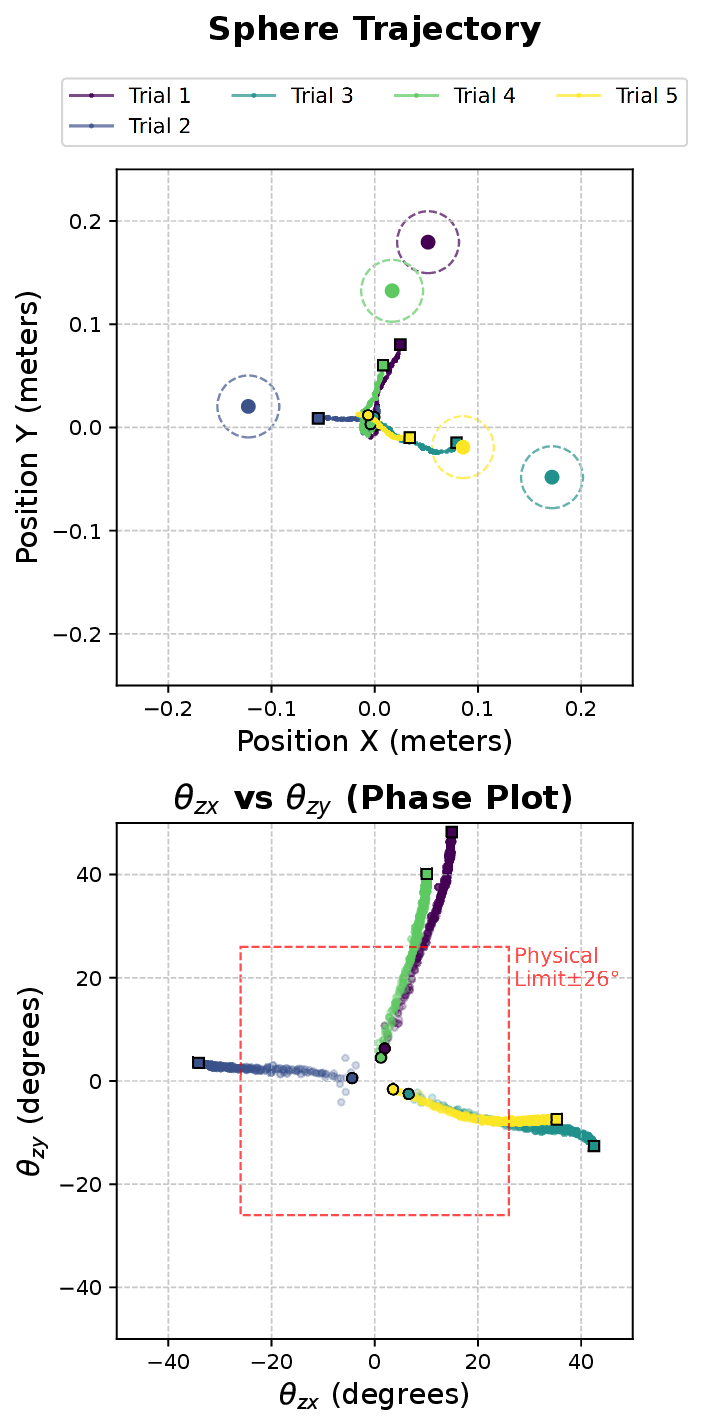} 
        \caption{Sphere, $\alpha = 0.0$}
        \label{fig:b_sphere}
    \end{subfigure}
    \hfill
    \begin{subfigure}[b]{0.23\textwidth}
        \centering
        \includegraphics[width=\textwidth]{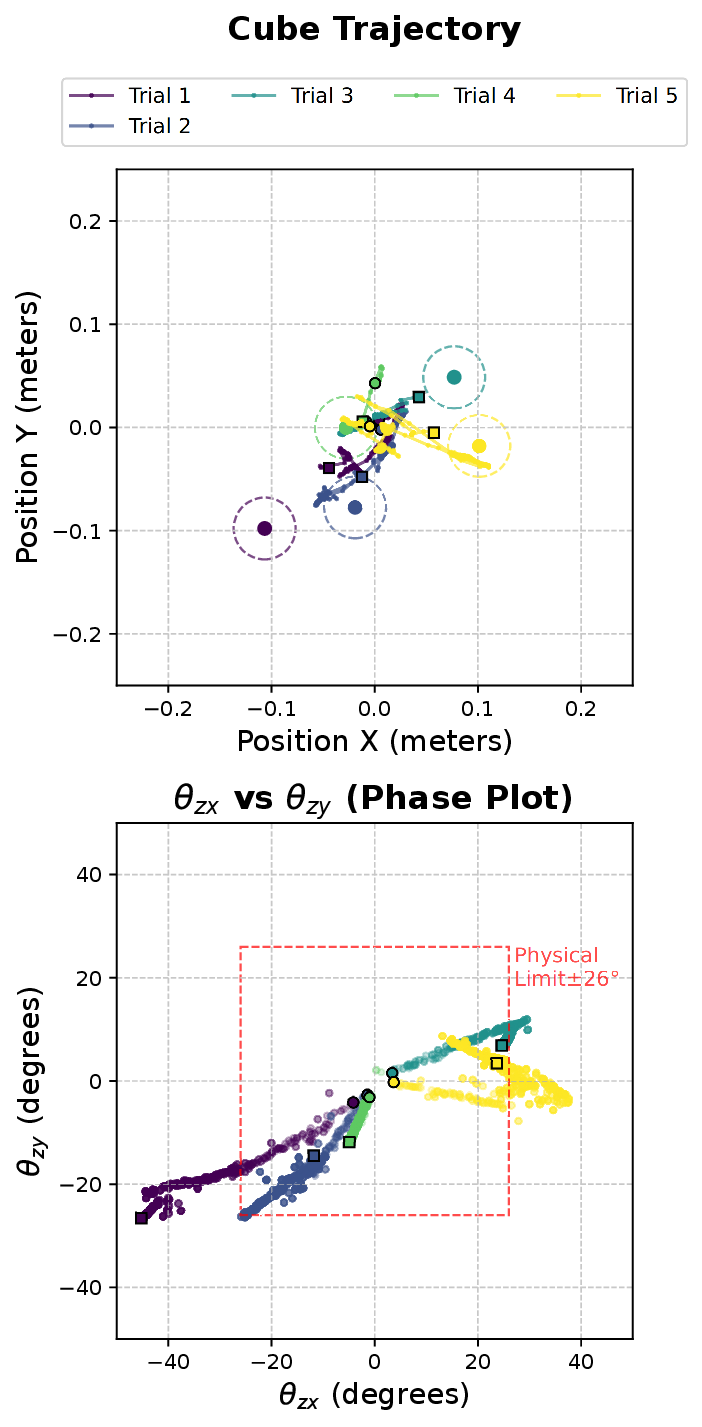} 
        \caption{Cube, $\alpha = 0.1$}
        \label{fig:b_cube}
    \end{subfigure}
    \hfill
    \begin{subfigure}[b]{0.23\textwidth}
        \centering
        \includegraphics[width=\textwidth]{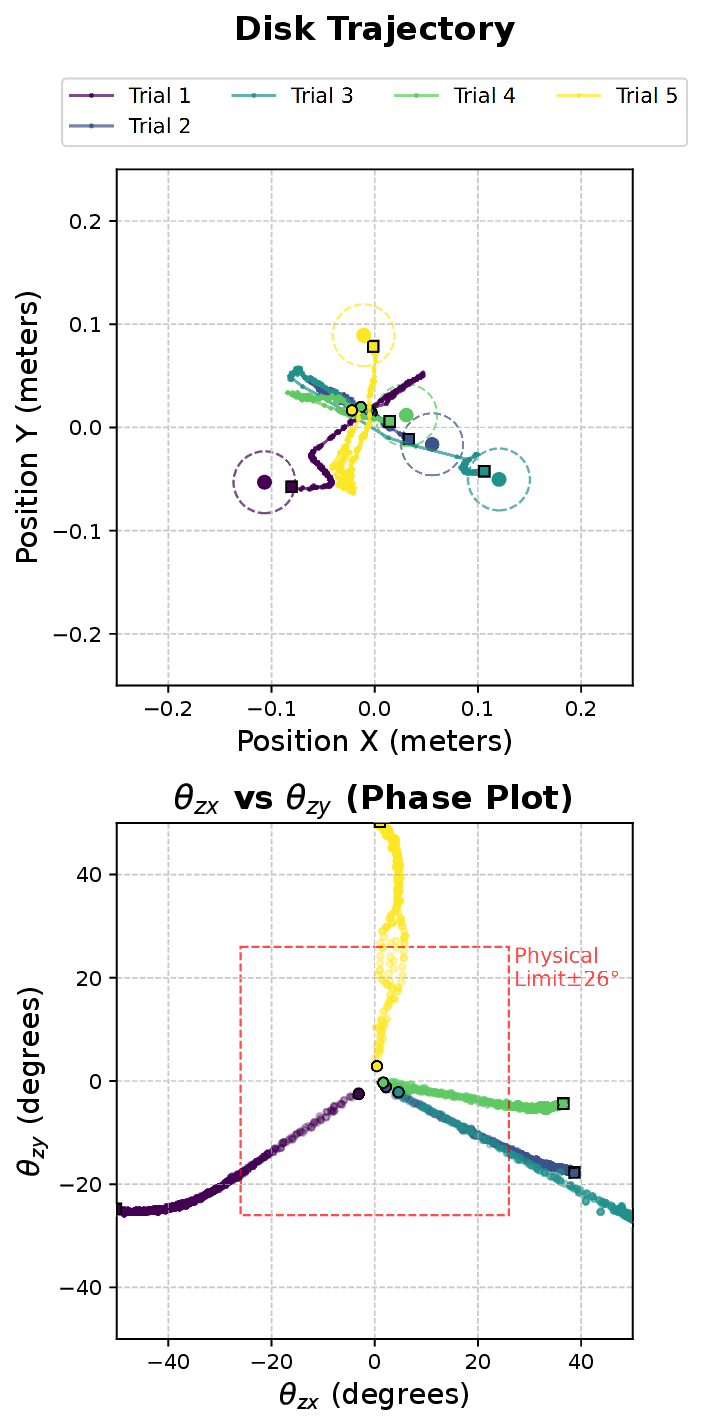} 
        \caption{Disk, $\alpha = 0.05$}
        \label{fig:b_disk}
    \end{subfigure}
    \hfill
    \begin{subfigure}[b]{0.23\textwidth}
        \centering
        \includegraphics[width=\textwidth]{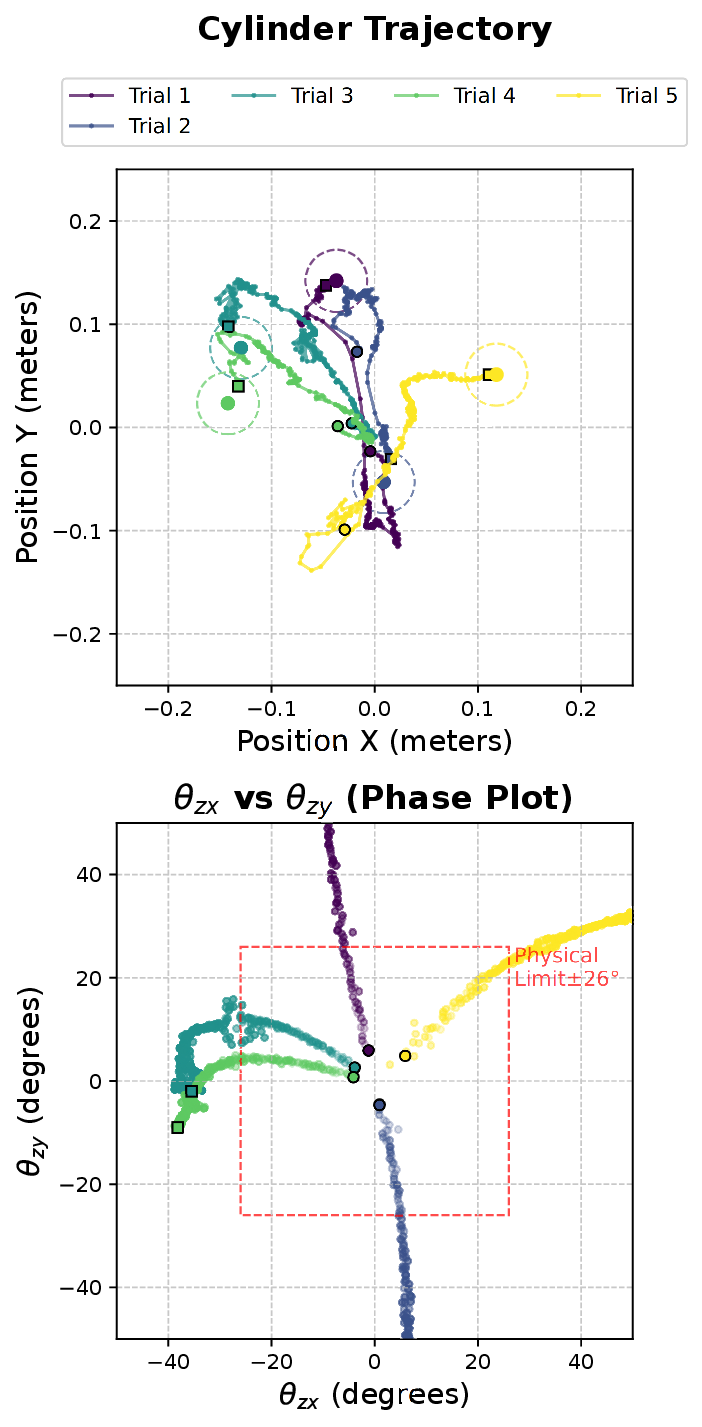} 
        \caption{Cylinder, $\alpha = 0.1$}
        \label{fig:b_apple}
    \end{subfigure}
    \hfill
    \begin{subfigure}[b]{0.23\textwidth}
        \centering
        \includegraphics[width=\textwidth]{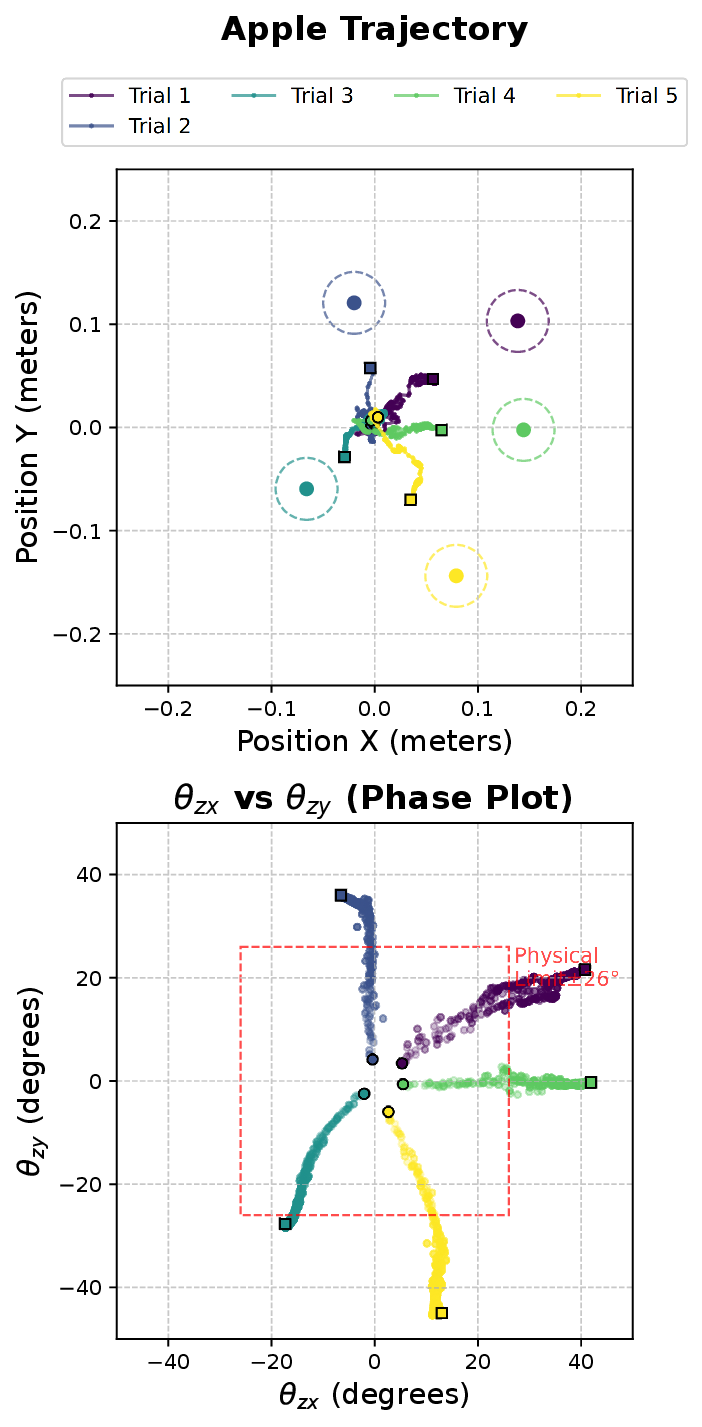} 
        \caption{Apple, $\alpha = 0.05$}
        \label{fig:b_egg}
    \end{subfigure}
    \hfill
    \begin{subfigure}[b]{0.23\textwidth}
        \centering
        \includegraphics[width=\textwidth]{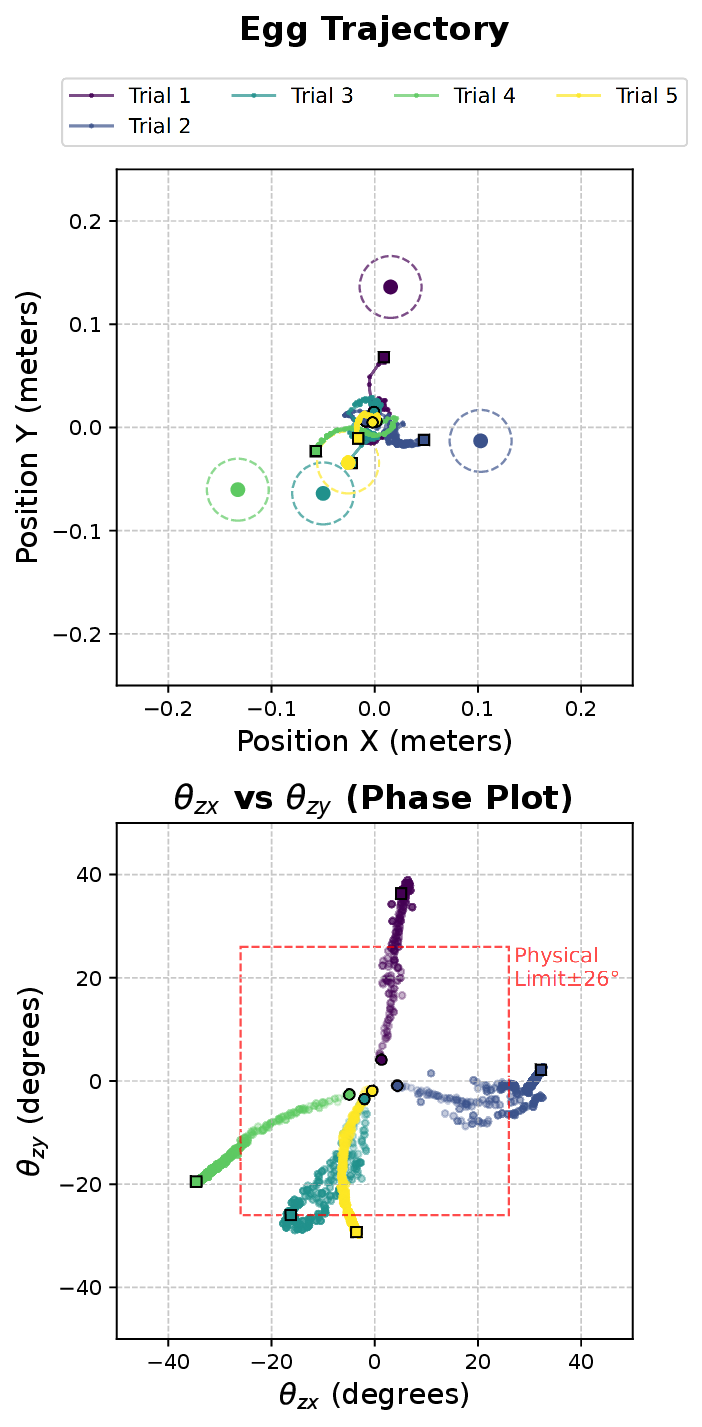} 
        \caption{Egg, $\alpha = 0.01$}
        \label{fig:b_random}
    \end{subfigure}
    \hfill
    \begin{subfigure}[b]{0.23\textwidth}
        \centering
        \includegraphics[width=\textwidth]{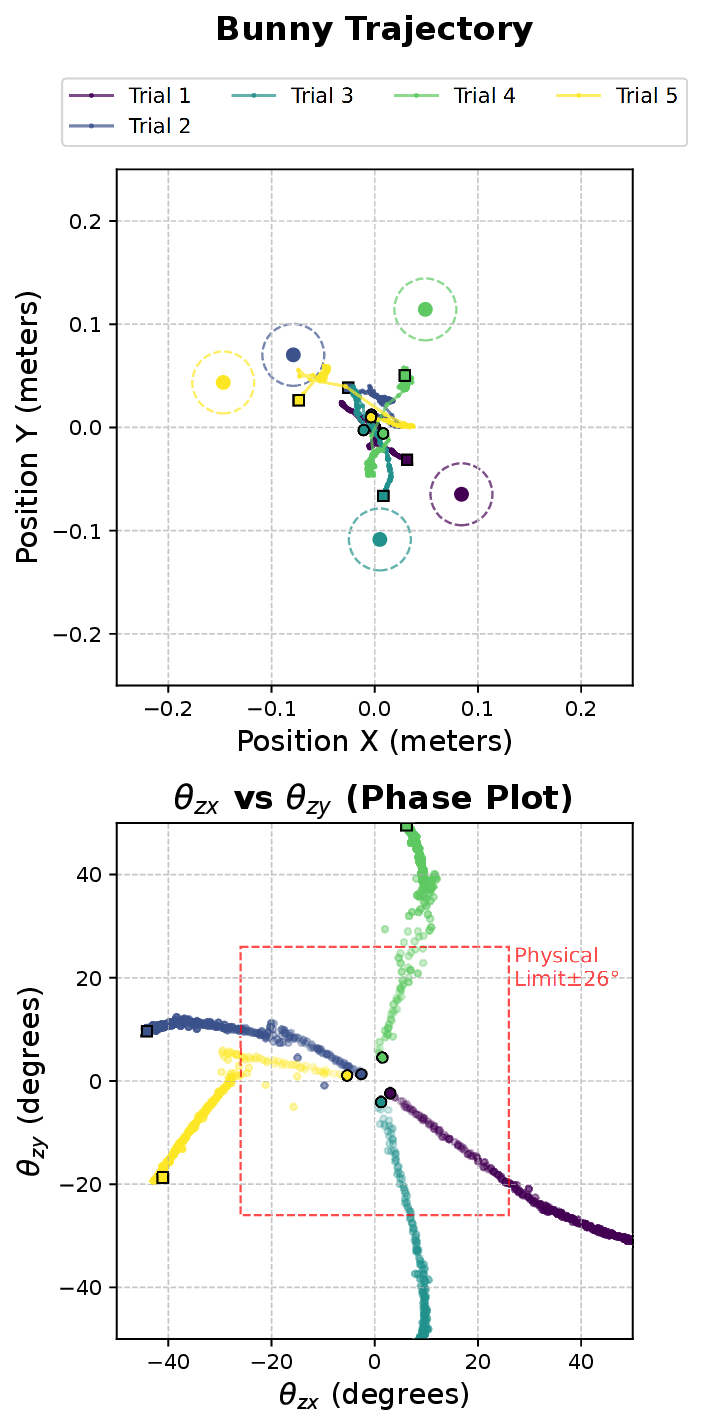} 
        \caption{Bunny, $\alpha = 0.05$}
        \label{fig:b_egg}
    \end{subfigure}
    \hfill
    \begin{subfigure}[b]{0.23\textwidth}
        \centering
        \includegraphics[width=\textwidth]{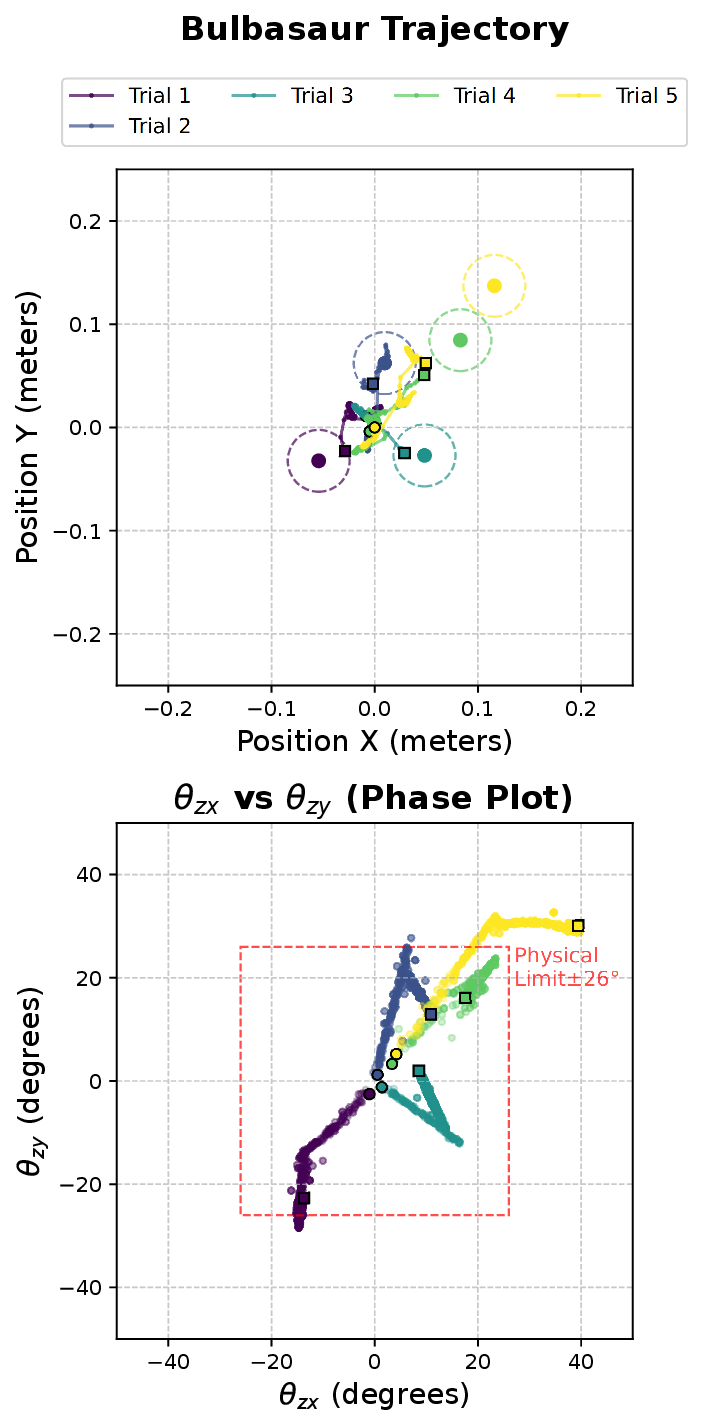} 
        \caption{Bulbasaur, $\alpha = 0.05$}
        \label{fig:b_random}
    \end{subfigure}
         
    \caption{Target reaching on the soft surface with various objects on the hardware system. The top line plot shows the trajectory taken by the objects, and the bottom phase plot shows the controller outputs $\theta_{zx}$ and $\theta_{zy}$. The red square represents the maximum angle values the system can take because of the physical range limit of the actuators $[-0.25, 0.25]$-meters.}
    \label{fig:trajectory_hardware}
\end{figure*}

Taget reaching on the hardware system is carried with five trials for each object. where each object is placed in the center of the fabric and an random desire position is given to the system. As seen in section \ref{e_v_m} both the controller struggles at the out edge of the fabric because of the curves of the fabric so the desire position is given with uniform probability between $[-0.2, 0.2]$-meter area on the fabric and trajectory followed by the object noted. Eight different object of different geometry, weight and texture used (table \ref{tab:objects}). Manhattan control policy with runtime of 1 minute and noise amplitude $\alpha =0.05$ is used for all the experiments as it give high performance, figure \ref{fig:trajectory_hardware}. For each object top line plot shows the path taken by the objects to reach the desired position (circular dashed region), bottom plot shows the phase plot $\theta_zx$ vs $\theta_zy$ return by the controller. Although the controller return an large angle output but the system can only facilitate angles between $[-26^\circ, +26^\circ]$ (red doted square) due to maximum limit of the actuator $[-0.25, 0.25]$. These angle are bounded when passed to the system. Based on the object characteristic the trajectories can be skewed. 

\section{Discussion}

The findings of this study demonstrate that a simplified linear PID-based feedback controller, enhanced by a geometric transformation of tilt angles, can robustly manipulate heterogeneous objects on the soft MANTA-RAY surface. By mapping a low-dimensional (one- or two-axis) tilt command to the surface’s actuators, the controller circumvents extensive training typically required by reinforcement learning-based methods, making the approach far more deployable in practice.

Comparisons between the Manhattan and Euclidean controllers underscore the superior performance of the Manhattan-based approach, which consistently demonstrated higher overall success rates and effective handling of target positions near the surface corners. This improvement arises from the independent axis control from Manhattan distance, which allows L-shape movements, allowing more precise shaping of the fabric to guide objects towards the desired position. Introducing a small, controlled noise term into actuator commands further enhances manipulation by preventing objects from becoming stuck in local fabric indentations.

Nonetheless, limitations remain. Near the periphery of the fabric, actuator travel constraints (±0.25 m) restrict tilt angles to ±26°, reducing the efficacy of both controllers at the edges. Additionally, inherent nonlinearities in fabric deformation may introduce subtle positioning errors that become more pronounced at lower control frequencies or with heavier objects. MANTA-RAY haively rely on single camera based visual position tracking of an object, limited by $12Hz$, putting constraints on the maximum attainable control frequency and singular points of failure in the systems. Investigating decentralized sensing \cite{dacre2025neuralcellularautomatadecentralized} approaches will significantly improve the control and edge cases.  

Future work may focus on adaptive tuning of PID parameters to improve corner and boundary performance, along with exploring larger-scale arrays of MANTA-RAY modules for distributed control approaches. Investigations into hybrid (model-based plus data-driven) control schemes could further refine the soft surface’s controllability, expanding potential applications in industrial logistics and automated handling of delicate, irregularly shaped goods.

\section{Conclusion}

This work demonstrates that a linear PID-based controller, augmented with a geometric transformation for tilt-angle mapping, provides an effective, generalized and scalable solution for manipulating heterogeneous objects on a soft robotic surface. By minimizing the need for extensive training and reducing hardware complexity to just four actuators, the proposed approach addresses the core limitations of traditional high-density array systems or learning-based methods. Experiments show that the Manhattan-based control strategy achieves superior success rates 36\% more compared to its Euclidean counterpart, especially near the corners of the manipulation surface, thus extending the functional workspace. Moreover, consistent handling of diverse and fragile items like eggs and apples underscores the adaptability of this method, supported by controlled noise that mitigates objects sticking in fabric indentations.

Despite the robustness shown, constraints remain regarding edge performance, primarily due to limited frame rate, actuator range and inherent non-linearities of the fabric. These findings encourage future studies focusing on adaptive PID gains and potentially hybrid controllers. Overall, the results validate a low-dimensional yet powerful control framework capable of addressing practical manipulation challenges across heterogeneous objects of different weights, sizes and textures, emphasizing the promise of soft robotic surfaces for broad industrial applications.
\bibliographystyle{unsrt}
\bibliography{ref}
\vspace{12pt}
\end{document}